\documentclass[lettersize,journal]{IEEEtran}
\usepackage{amsmath,amsfonts}
\usepackage{algorithmic}
\usepackage{algorithm}
\usepackage{array}
\usepackage[caption=false,font=normalsize,labelfont=sf,textfont=sf]{subfig}
\usepackage{textcomp}
\usepackage{stfloats}
\usepackage{url}
\usepackage{verbatim}
\usepackage{graphicx}
\usepackage{cite}
\usepackage{xcolor}
\usepackage{tikz}
\usetikzlibrary{shapes, arrows, positioning}
 \usepackage{booktabs} 
\usepackage{multirow}
\usepackage{makecell}
\usepackage{hyperref}
\usepackage[normalem]{ulem}
\usepackage[all]{nowidow}
\begin{document}

\title{A Novel Evolutionary Method for Automated Skull-Face Overlay in Computer-Aided Craniofacial Superimposition}


\author{\IEEEauthorblockN{Práxedes Martínez-Moreno*,
Andrea Valsecchi*, Pablo Mesejo, Pilar Navarro-Ramírez, Valentino Lugli, and Sergio Damas}
\thanks{Práxedes Martínez-Moreno and Pablo Mesejo with the Department of Computer Science and Artificial Intelligence, University of Granada, Spain and the Andalusian Research Institute in Data Science and Computational Intelligence, University of Granada, Granada, Spain (e-mail: praxedesmm@ugr.es; pmesejo@ugr.es).}
\thanks{Andrea Valsecchi, Pilar Navarro-Ramírez, and Valentino Lugli with Panacea Cooperative Research S. Coop., Ponferrada, Spain (e-mail: valsecchi.andrea@gmail.com; pilarnavar98@gmail.com; valentinolugli@gmail.com).}
\thanks{Sergio Damas with the Department of Software Engineering, University of Granada, Spain and the Andalusian Research Institute in Data Science and Computational Intelligence, University of Granada, Granada, Spain (e-mail: sdamas@ugr.es).}
\thanks{* These authors contributed equally to this work.}}



\maketitle

\vspace{-1em}
\begin{center}
\textit{This work has been submitted to the IEEE for possible publication. Copyright may be transferred without notice, after which this version may no longer be accessible.}
\end{center}

\begin{abstract}
Craniofacial Superimposition is a forensic technique for identifying skeletal remains by comparing a post-mortem skull with ante-mortem facial photographs. A critical step in this process is Skull-Face Overlay (SFO). This stage involves aligning a 3D skull model with a 2D facial image, typically guided by cranial and facial landmarks' correspondence. However, its accuracy is undermined by individual variability in soft-tissue thickness, introducing significant uncertainty into the overlay. This paper introduces Lilium, an automated evolutionary method to enhance the accuracy and robustness of SFO. Lilium explicitly models soft-tissue variability using a 3D cone-based representation whose parameters are optimized via a Differential Evolution algorithm. 
The method enforces anatomical, morphological, and photographic plausibility through a combination of constraints: landmark matching, camera parameter consistency, head pose alignment, skull containment within facial boundaries, and region parallelism. 
This emulation of the usual forensic practitioners' approach leads Lilium to outperform the state-of-the-art method in terms of both accuracy and robustness.
\end{abstract}


\section{Introduction}
\begin{figure}[h]
    \centering
    \includegraphics[width=\linewidth]{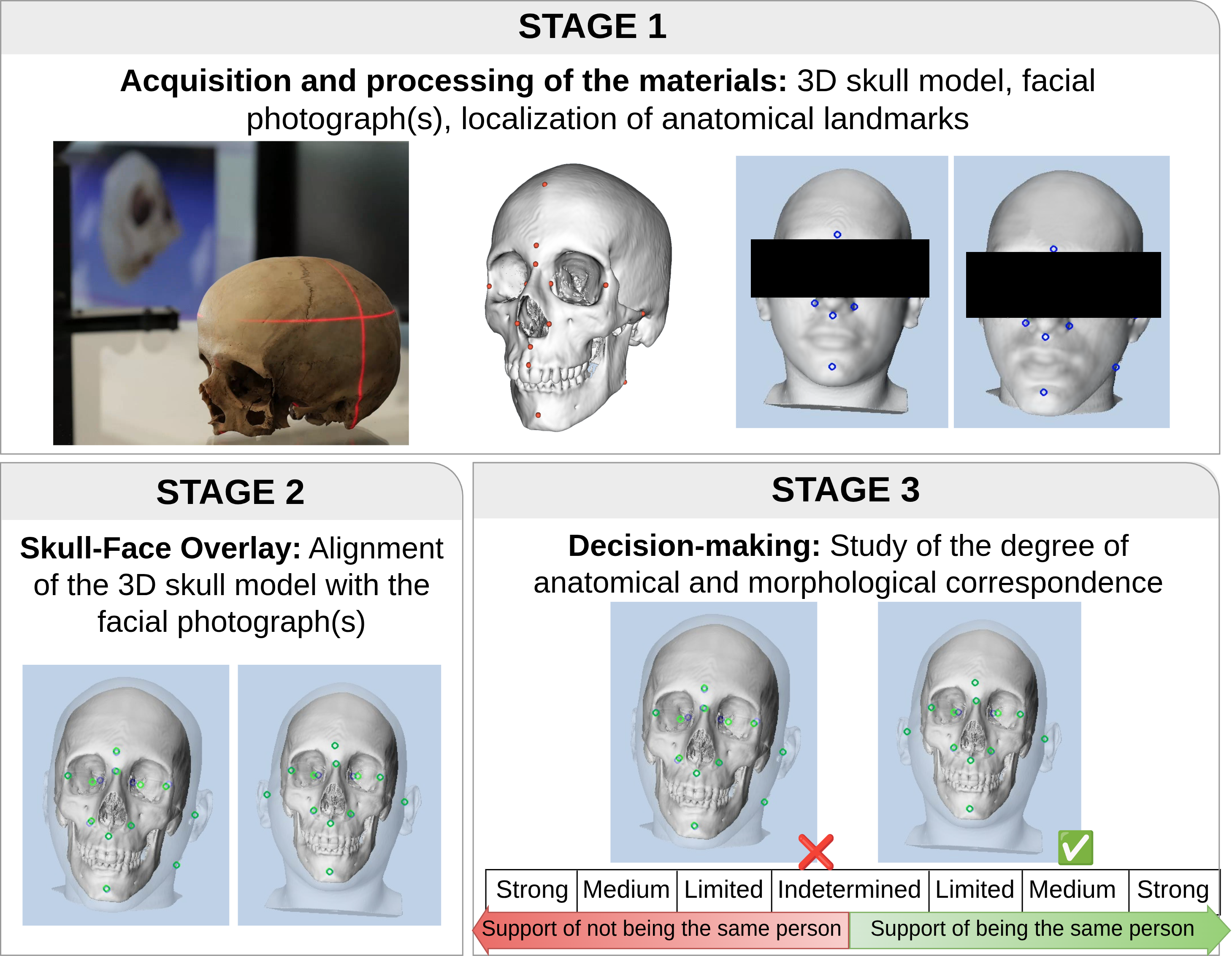}
    
    \caption{Overview of the CFS process comprising three main stages.}
    \label{fig:sfo}
\end{figure}

\IEEEPARstart{C}{raniofacial} Superimposition (CFS) is a forensic technique used to aid identification when dealing with skeletal remains \cite{Yoshino12}. 
CFS has been applied to support the identification in war, crime, and nature casualties, among others \cite{blau2009handbook}. In such scenarios, the application of well-known identification techniques (such as fingerprints or DNA) is often unfeasible due to the lack of either ante-mortem or post-mortem crucial information \cite{Leggio2025}. The technique assists forensic practitioners by comparing one or more ante-mortem facial photographs with a recovered post-mortem skull. In order to achieve this, the projection of the latter onto the available facial images is performed.
%
%
Three stages are distinguished in the CFS process \cite{Damas11CSUR, HUETE2015267} (represented in Figure~\ref{fig:sfo}): 

\begin{enumerate}
    \item Acquisition and processing of the post-mortem skull and ante-mortem facial photograph(s), together with the location of anatomical landmarks. These landmarks fall into two categories: craniometric landmarks (hereafter ``cranial landmarks''), which are anatomically defined points on the skull; and cephalometric landmarks (hereafter ``facial landmarks''), which are corresponding points located on the face. The interested reader is referred to \cite{Burns07} for the location and name of the cranial and facial landmarks typically used in CFS.
    The correspondence between a cranial and a facial landmark is not exact due to the presence of soft tissue, which surrounds and protects the body structures and organs. 
    \item Skull-Face Overlay (SFO), which aims to achieve the most accurate alignment of a 3D skull model with a facial photograph, often by matching corresponding cranial and facial landmarks. When multiple photographs are available, separate SFO processes are~conducted. 
    \item Decision-making, which is performed by the practitioner assessing the degree of anatomical and morphological correspondence based on the SFO to determine whether the skull and facial photograph(s) belong to the same individual.
\end{enumerate}

Despite its extensive applicability and significant value, the CFS technique remains intricate and challenging to perform \cite{Burns07}.
Historically, one of the main drawbacks of this technique has been its reliance on non-standardized methods \cite{Damas15FSI, Ibanez11TFS, ibanez2016meprocs}. 
The MEPROCS European project\footnote{\url{https://cordis.europa.eu/project/id/285624/es}} was developed to overcome this challenge, with its results and best practice guidelines presented in \cite{DamasHandbook2020}.
Despite these efforts, the three stages of CFS are still performed manually in most forensic laboratories. The SFO stage, in particular, remains highly complex and time-consuming, often requiring hours of trial and error to achieve an optimal alignment between the skull and facial photograph~\cite{Damas11CSUR,Ibanez11TFS}.

Our work specifically contributes to the second stage of the CFS technique: the SFO process. This task can be framed as a 3D-2D Image Registration (IR) problem, where the goal is to align two different objects: a 3D skull model with a 2D facial photograph \cite{Campomanes15TIFS}. IR aims to identify the transformation (e.g., rotation and translation, among others) that superimposes two or more images captured under distinct conditions \cite{ZITOVA2003977}. In the CFS context, IR is a challenging optimization problem. Evolutionary Algorithms (EAs) have emerged as a promising solution due to their global optimization capabilities, enabling robust searches even in complex and ill-defined problems 
\cite{damas2011a, santamaria2011, campomanes2014computer, Campomanes15TIFS}. The SFO process thus involves determining the ideal skull pose (its position and orientation in three-dimensional space) and camera settings to ensure that the projected cranial landmarks align with their facial counterparts, 
accounting for the soft tissue between every pair of landmarks.

SFO also faces difficulties due to subjectivity and incomplete information. Accurately pinpointing certain landmarks can be highly subjective \cite{Campomanes15IJLM, Cummaudo13IJLM}. Several factors further complicate this process, such as variations in the mandible articulation, the pose of the subject in the photo, and occlusions from hair, beards, glasses, clothing, or overall image quality. Regarding the skull, structural damage or missing parts can hinder the accurate location of landmarks. Furthermore, the specific thickness of the soft tissue separating homologous cranial and facial landmarks for an individual remains unknown. 
Although general estimates based on population studies are available \cite{Stephan2008}, and some studies have attempted to predict soft-tissue thickness using skull morphology and biological profile \cite{guyomarc14}, significant uncertainties continue to exist \cite{VALSECCHI2018}.

The current state-of-the-art method in automated SFO was introduced in \cite{VALSECCHI2018}, namely POSEST-SFO. It is recognized for its high performance and accuracy. However, unlike earlier approaches \cite{Campomanes15TIFS}, this method does not directly model important sources of uncertainty, such as the aforementioned estimation of soft-tissue thickness. Instead, this method requires (as input) 3D vectors connecting cranial and facial landmarks for the individual under study. This is unrealistic in actual caseworks. Moreover, there is only one of such vectorial studies \cite{NAVARRO2026112802}.

In this paper, we introduce Lilium, an automated evolutionary method \cite{eiben2015introduction} for the SFO stage of CFS. Lilium explicitly handles uncertainty regarding soft tissue to improve the alignment of cranial and facial landmarks. 
The method considers for the first time the anatomical and morphological correspondence between the skull and face, as well as the camera settings and head pose, to ensure that the overlays are plausible.
Bilateral landmarks are optimized jointly to enforce consistency and prevent asymmetrical distortions. 
To quantitatively validate Lilium, we simulate controlled SFO cases providing exact ground-truth data, enabling rigorous comparisons. These innovations allow for robust SFOs across various scenarios with increasing levels of difficulty. 
In essence, our contribution makes the automatic overlapping process more analogous to the methodology employed by a forensic anthropologist, and allows us to yield both state-of-the-art and more explainable results in this challenging~problem.

\section{Our Proposal: Lilium}
\label{sec:lilium}

\subsection{From Camera Calibration and 3D Pose Estimation to SFO}
\label{sec:bg}

Registering a 3D skull model with a 2D facial photograph
involves determining the relative position and orientation between the head and the camera. This task hinges on both the pose of the camera and its intrinsic parameters (e.g., focal length, principal point, and lens distortion).
``Camera calibration'' refers to the process of identifying these parameters typically achieved by capturing images of an object with known dimensions or multiple views of an unknown object. The ``Perspective-n-Point'' (\textit{PnP}) problem, instead, entails computing the pose of the camera based on a set of $n$ known 3D points and their corresponding projections onto the image \cite{hartley2003multiple}. In many cases, image formation is approximated by the pinhole model, reducing complexity to just the focal length and thereby merging calibration and pose estimation into what is known as the \textit{PnP+f} problem, where \textit{f} stands for focal \cite{VALSECCHI2018}.

Existing methods, such as RCGA \cite{Ibanez11TFS} and RCGA-C \cite{Campomanes15TIFS}, leverage EAs to tackle this challenge through iterative and stochastic optimization. In contrast, an algebraic approach was introduced that finds an optimal solution by solving polynomial equations for $n\geq4$ \cite{bujnak2008general}. These equations relate the distances between the $n$ points before and after their projection. This technique was later integrated into a RANSAC framework \cite{fischler1981random} with an additional local optimization step, resulting in the POSEST algorithm \cite{lourakis2013model}. Extending these developments, POSEST-SFO resolves the SFO problem taking into account the presence of the soft tissue, representing the state-of-the-art automatic SFO method. However, it does not model the uncertainty related to soft tissue. Rather, the vectors indicating its direction at each landmark are provided as input, and their lengths are fixed at the average soft-tissue depths reported in \cite{Stephan2008}. 
Once this adjustment is made, POSEST is applied for the \textit{PnP+f} computation \cite{VALSECCHI2018}. 

\subsection{Soft-Tissue Modeling with 3D Cones}
\label{sec:cones}

\begin{figure*}[h]
    \centering
    \includegraphics[width=\textwidth]{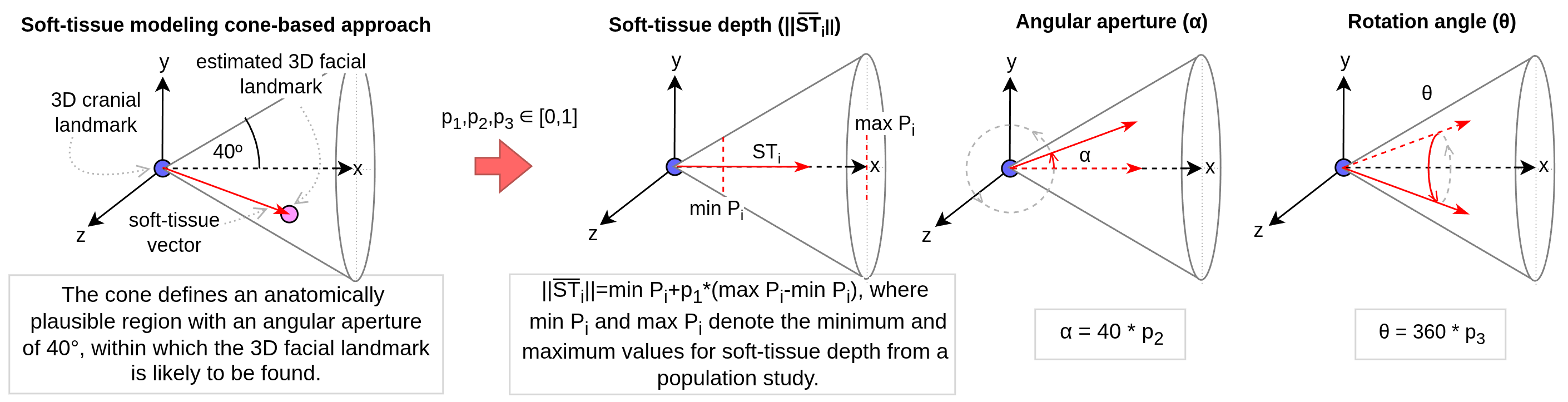}
    \caption{Cone-based soft-tissue modeling approach for 3D facial landmark estimation.
    The volume defines an anatomically plausible conical search region with a $40^\circ$ angular aperture, originating from a cranial landmark (left). The soft-tissue depth ($||\vec{ST}_i||$) is calculated as a linear interpolation between population-based minimum ($minP_i$) and maximum ($maxP_i$) values using the parameter $p_1$ (second from left). 
    The angular aperture ($\alpha$) of the cone is scaled by $p_2$, and the rotation angle ($\theta$) about the cranial vector is controlled by $p_3$, which allows rotation across the entire $360^\circ$ range (third and fourth from left, respectively). 
    Together, these parameters define a constrained 3D region where the facial landmark is likely to be found.}
    \label{fig:sampling}
\end{figure*}

In the SFO process, RCGA-C models soft-tissue uncertainty using fuzzy sets shaped like a 3D cone. Our method, Lilium, adopts a cone-based representation that builds on the core idea introduced in RCGA-C \cite{Campomanes15TIFS}, but simplifies it significantly. Instead of relying on fuzzy membership functions to define soft-tissue uncertainty, we use a geometric 3D cone whose key elements are defined as~follows:
\begin{itemize}
    \item The cone vertex is positioned at the 3D cranial~landmark.
    \item The axis of the cone is approximated by the expected direction of soft-tissue growth at the cranial landmark. This rough direction is derived from anatomical or empirical data gathered by forensic practitioners \cite{NAVARRO2026112802, Stephan2008}.
    \item The height of the cone corresponds to the landmark maximum soft-tissue depth derived from a statistical population study we conducted for this work. To ensure a broader sample, this study involved our core dataset (Section~\ref{sec:data_generation}) supplemented by data from six additional CBCT scans\footnote{These six subjects were excluded from our experiments because they lacked the complete forehead region, which is essential for our~proposal.}.
    \item The volume of the cone represents the spatial region where the corresponding 3D facial landmark is expected to be located. 
\end{itemize}

In our approach, the 3D facial landmarks are sampled within the cones using three parameters that define a soft-tissue vector ($\vec{ST_i}$). This vector represents the displacement between a 3D cranial landmark ($C_i$) and the corresponding 3D facial landmark ($F_i$), expressed as $\vec{ST_i} = F_i - C_i$. Figure~\ref{fig:sampling} illustrates the overall sampling process.
The parameters used to define a soft-tissue vector are:
    \begin{itemize}
        \item $p1$: Adjusts the length of the vector originating from the cranial landmark, determining the soft-tissue depth. 
        \item $p2$: Controls the angular aperture 
        from the axis of the cone, capturing natural variation in the direction of soft-tissue growth. In our study, this angle ranges from $0^\circ$ to $40^\circ$. This choice is supported by a preliminary analysis, which indicated that deviations within this range yield anatomically consistent landmark placements. Moreover, this aligns with prior work such as RCGA-C-45 \cite{Campomanes15TIFS} that demonstrated superior performance. 

        \item $p3$: Modifies the rotation angle about the axis of the cone, introducing circumferential variation 
        in the soft-tissue vector. It can take any value in the range $[0^\circ, 360^\circ]$.
    \end{itemize}

These three parameters take values within $[0,1]$. The proposed representation thus provides a biologically plausible region for each 3D facial landmark, reduces model complexity, and offers a more transparent and interpretable~alternative.

\subsection{Evolutionary Optimization of Soft-Tissue Modeling}
\label{sec:opt}

\begin{figure}[h]
  \centering
  \includegraphics[width=0.49\textwidth]{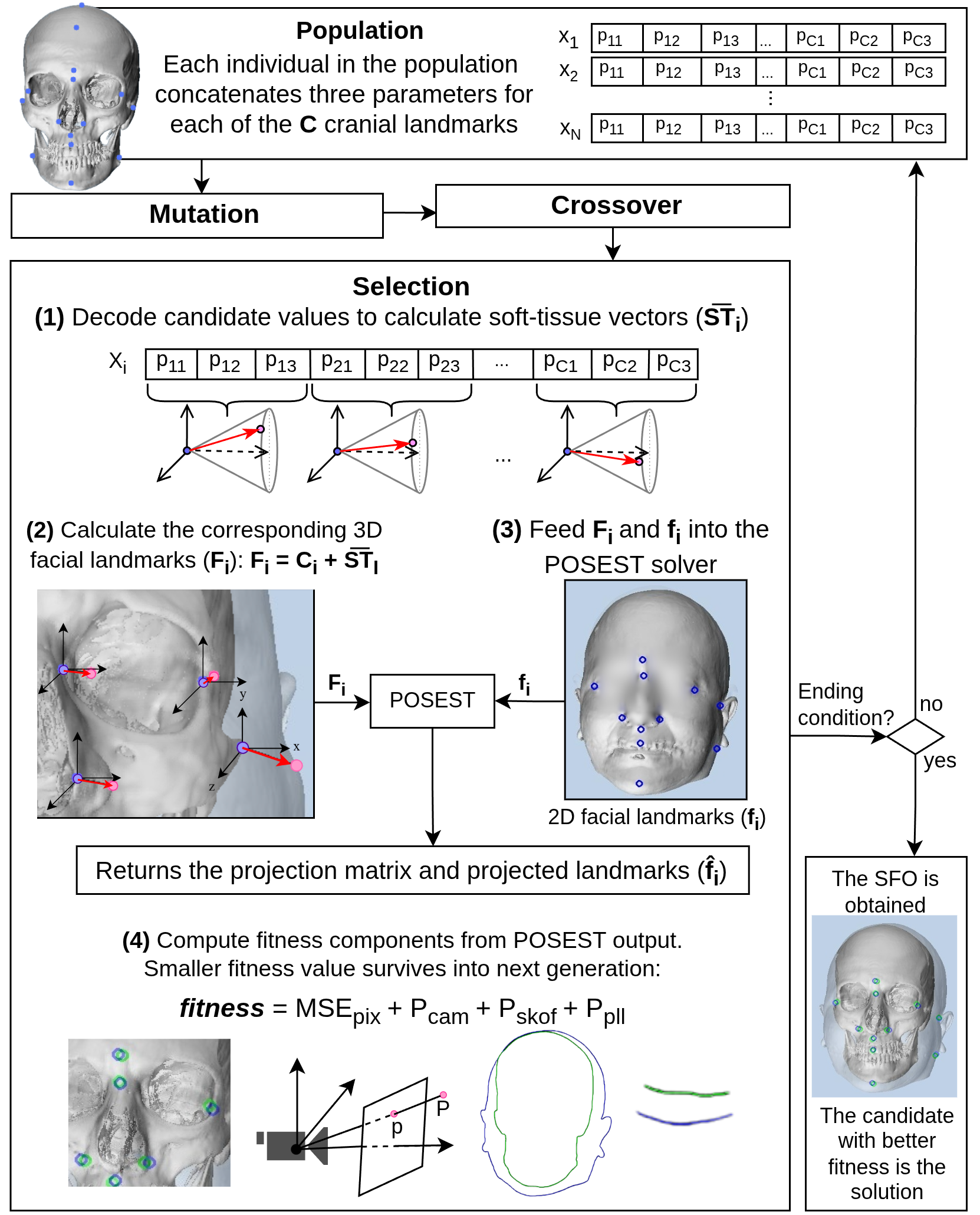}
  \caption{This figure illustrates the Lilium pipeline. The process begins with the random initialization of a solution population, which is then iteratively refined by the DE algorithm through mutation, crossover, and selection. A key element in this refinement is Lilium's powerful composite fitness function (see Section~\ref{sec:opt}). This iterative process continues until the defined stopping criterion is met, at which point the vector with the best fitness score represents the optimal SFO solution.}
  \label{fig:de_flow_complete}
\end{figure}

In order to solve the soft-tissue modeling problem within the anatomically constrained 3D cone space, we employ the Differential Evolution (DE) algorithm, which is a population‐based, stochastic optimizer well suited for non‐linear, non‐convex search spaces with mixed continuous parameters. 
Specifically, we use the rand/1/exp variant, as its inherent tunable balance between exploration (through mutation) and exploitation (through crossover) is crucial for navigating the intricate landscape of soft-tissue deformation \cite{storn1997differential}. 
While RCGA-C optimizes twelve parameters of the 3D-to-2D geometric transformation that projects the skull onto the photograph, Lilium optimizes three parameters ($p_1$, $p_2$, and $p_3$) that determine the estimated 3D location of each facial landmark.
Then, POSEST is used to compute the pose of the skull based on these estimated 3D landmarks and those from the facial image. The overall DE workflow is illustrated in Figure~\ref{fig:de_flow_complete}.


The genotype for each individual in the DE population is defined as a vector $\mathbf{x}$ that concatenates the three cone parameters for each of the $C$ cranial landmarks. In the case of paired bilateral landmarks, right and left counterparts tend to exhibit a strong correlation in thickness \cite{CHUNG2015132}. To model this anatomical correspondence, symmetry is encouraged through a hybrid strategy. For each bilateral pair, the three cone parameters of one landmark are generated via the usual sampling process, while those of its symmetrical counterpart are computed as a weighted combination: 90\% drawn from the sampled values of the first landmark and 10\% drawn from an independent sample. This approach preserves the expected correlation between bilateral landmarks, while allowing for sufficient variability to reproduce realistic anatomical differences. Note that the global transformation of RCGA-C cannot capture this level of detail. Therefore, each candidate solution in the DE population takes the form
\(
\mathbf{x} = \bigl[\, p_{11},\, p_{12},\, p_{13},\, \ldots,\, p_{C1},\, p_{C2},\, p_{C3} \,\bigr],
\)
where $p_{i1}$, $p_{i2}$, and $p_{i3}$ modify the depth, angular aperture, and rotation angle for each cone, respectively.
At the beginning, this population is initialized with random values drawn from the range $[0,1]$. In each generation, every member of the population gives rise to a trial vector through a two-step process: first, three distinct individuals are chosen at random and their differences are linearly combined to form a mutant vector; then, the exponential crossover process combines the mutant with the original individual at a given probability.
A composite fitness score is then evaluated for every trial alongside its parent, and the vector with better fitness survives into the next generation.

In order to assess the quality of a DE individual $\mathbf{x}$, we first follow this pipeline (also illustrated in Figure~\ref{fig:de_flow_complete}): For each cranial landmark $C_i$, we decode its soft-tissue vector $\vec{ST_i}$ from its parameters ($p_{i1}$, $p_{i2}$, and $p_{i3}$) inside the 3D cone. Subsequently, these soft-tissue vectors are used to generate 3D facial landmarks by computing $F_i = C_i + \vec{ST_i}$ for all $i$. Finally, for pose estimation, the 3D-2D correspondences $(F_i,\,f_i)$ are fed into the POSEST solver, where $f_i$ is the matching 2D landmark previously pinpointed in the facial image. POSEST thus returns the projection matrix $P$ (including camera intrinsic and extrinsic parameters) and the projected points $\hat{f_i} = P(F_i)$.

The quality of a candidate is thus defined by combining: 
(1) the accuracy of the matching between every pair of projected landmark $\hat{f_i}$ and its corresponding facial landmark in the photograph $f_i$,
and (2) anatomical, morphological, and photographic plausibility penalties.  
Lower fitness indicates a better SFO.  
The fitness of a candidate for a specific facial photograph is calculated as 
\( 
fitness = \text{MSE}_{\text{pix}} + \text{P}_{\text{cam}} + \text{P}_{\text{skof}} + \text{P}_{\text{pll}},
\)
where:
\begin{itemize}
    \item $\text{MSE}_{\text{pix}}$ is the Mean Squared Error (MSE) in pixels between the projected 3D facial landmarks and their corresponding 2D points in the photograph:
    $\mathrm{MSE}_{\mathrm{pix}}=\frac{1}{N}\sum_{i}\|\hat f_i - f_i\|^2$.

    \item $\text{P}_{\text{cam}}$ is a penalty related to the camera intrinsic and extrinsic parameters,
    as well as the pose difference relative to the one shown in the photograph ($\Delta\beta$). This term is outlined in Section~\ref{sec:camera_params}.

    \item $\text{P}_{\text{skof}}$ is a penalty applied when the skull projection extends beyond the boundaries of the face (detailed in Section~\ref{sec:skull_face}). The name ``skof'' is a shorthand for \textit{skull-outside-face}. 

    \item $\text{P}_{\text{pll}}$ is a term that quantifies the lack of parallelism between the skull and facial chin-jaw or forehead regions (explained in Section~\ref{sec:chin_forehead}).
\end{itemize}
 
The cycle of mutation, crossover, and selection is repeated across the population until the stopping criterion is met, ensuring exploration while preserving high‐quality solutions.
The best individual in terms of fitness yields the optimized soft‐tissue parameters, which lead to the final SFO. 
Note that the goal is to achieve SFOs in which the penalty terms are zero or close to zero, indicating no violations of anatomical, morphological, or geometric plausibility. Therefore, while fitness serves as a guide during optimization, its actual value is secondary to ensuring that all constraints are satisfied.

The stopping criterion was determined through a preliminary empirical analysis of convergence behavior. For all compared methods (see Section~\ref{sec:settings}), the median relative improvement in best-so-far fitness rapidly decreased during early evolution and plateaued around 500 generations, with only marginal gains thereafter. Approximately 500 seconds captured nearly all meaningful improvements while avoiding unnecessary computational cost. To ensure sufficient exploration without favoring faster methods, a maximum of 750 generations was also imposed. Execution was therefore terminated upon reaching either 500 seconds or 750 generations.

\subsection{Camera Parameters Criterion}
\label{sec:camera_params}

In order to guarantee that the estimated projection fulfills realistic photographic constraints, we introduce a camera-parameter penalty term, \( \text{P}_{\text{cam}} \). This term accounts for focal length (\( f_x \)), subject-to-camera distance (SCD), and the angular deviation between the inferred and reference head poses (\( \Delta\beta \)). 

The values of \( f_x \), SCD, and pose angles are obtained through a dual estimation strategy. On the one hand, independent estimates (referred to as ``a priori information'') are derived directly from the input image using machine learning techniques. For instance, SCD can be estimated using methods such as the CNN-based approach introduced by Bermejo et al.~\cite{BERMEJO2022118457}. Inference of head pose can be facilitated by employing methods similar to those outlined by Chun et al.~\cite{CHUN2025}. The focal length can be obtained either from the EXIF metadata of the image or via monocular camera calibration, such as the deep learning approach proposed by Workman et al.~\cite{WORKMAN2015}. Each estimated quantity is associated with an admissible interval that reflects its expected range of variation. These intervals are denoted as 
\( f_x \in [\mathit{ap}_{f_x,\min}, \mathit{ap}_{f_x,\max}] \) and 
\( \mathrm{SCD} \in [\mathit{ap}_{\mathrm{SCD},\min}, \mathit{ap}_{\mathrm{SCD},\max}] \),
while pose consistency is constrained by a maximum angular deviation \( \beta_{\mathrm{tol}} \).

In parallel, the same parameters are also extracted from the output of the POSEST solver. These values are evaluated against the defined admissible intervals. Any parameter deviating beyond its expected range is penalized within the fitness function.
The penalty term is defined as 
%
\begin{align*}
\text{P}_{\mathrm{cam}} 
&= \mathbf{1}_{\{f_x=0\}}\,C_\infty
  + \left[\max(0,\; \mathit{ap}_{f_x,\min} - f_x)\right]^2 \notag \\
&\quad + \left[\max(0,\; f_x - \mathit{ap}_{f_x,\max})\right]^2\\
&\quad+ \left[\max(0,\; \mathit{ap}_{\mathrm{SCD},\min} - \mathrm{SCD})\right]^2 \notag \\
&\quad + \left[\max(0,\; \mathrm{SCD} - \mathit{ap}_{\mathrm{SCD},\max})\right]^2\\
&\quad+ \left[\max(0,\; \Delta\beta - \beta_{\mathrm{tol}})\right]^2,
\label{eq:camera_penalty}
\end{align*}
%
where the term \( \mathbf{1}_{\{f_x=0\}}\,C_\infty \) introduces a large penalty if the estimated \( f_x \) collapses to zero, indicating an invalid projection. The squared penalty terms ensure that deviations from the permissible intervals for \( f_x \) and SCD are penalized in proportion to their magnitude. Similarly, if the angular difference \( \Delta\beta \) between the optimized and image-inferred pose exceeds the tolerance \( \beta_{\mathrm{tol}} \), the penalty increases quadratically.
In practice, if POSEST does not find a solution or \( f_x \) exceeds a predefined realistic limit, a significantly large penalty is applied. This realistic limit is based on real-world camera lens capabilities and is set to 10,000 in our study.
Finally, by incorporating \( \text{P}_{\text{cam}} \) into the overall fitness function, we ensure that the final alignment is consistent with physically feasible camera and head-pose parameters.

\subsection{Skull-Outside-Face Criterion}
\label{sec:skull_face}

In order to promote anatomically plausible alignments, we introduce the \(\text{P}_{\text{skof}}\) term, which penalizes configurations in which the projected skull contour extends beyond the visible facial boundary in the image. 

The computation proceeds as follows: first, the full facial contour is extracted from the input image, and the skull is projected onto the image plane using the projection matrix returned by POSEST. 
The projected skull region is then rasterized to produce a binary mask. The penalty is defined as the number of pixels in the skull mask that fall outside the facial mask. Formally, letting \( \mathcal{S} \) denote the set of pixels occupied by the projected skull and \( \mathcal{F} \) the facial region, the penalty is given by
\(
\text{P}_{\text{skof}} = 100 \times \left| \mathcal{S} \setminus \mathcal{F} \right|,
\)
where \( |\cdot| \) denotes the set cardinality, i.e., the number of pixels. This quantity is multiplied by a scaling factor of 100 to emphasize the impact of anatomically implausible configurations and ensures that they are strongly penalized in the optimization process. 
By incorporating \(\text{P}_{\text{skof}}\) into the overall fitness function, the method steers the optimization towards alignments in which the skull remains spatially enclosed by the facial region.

\subsection{Parallelism Criterion: Chin-Jaw and Forehead Curves}
\label{sec:chin_forehead}

\begin{figure}[h]
    \centering
    \includegraphics[width=\linewidth]{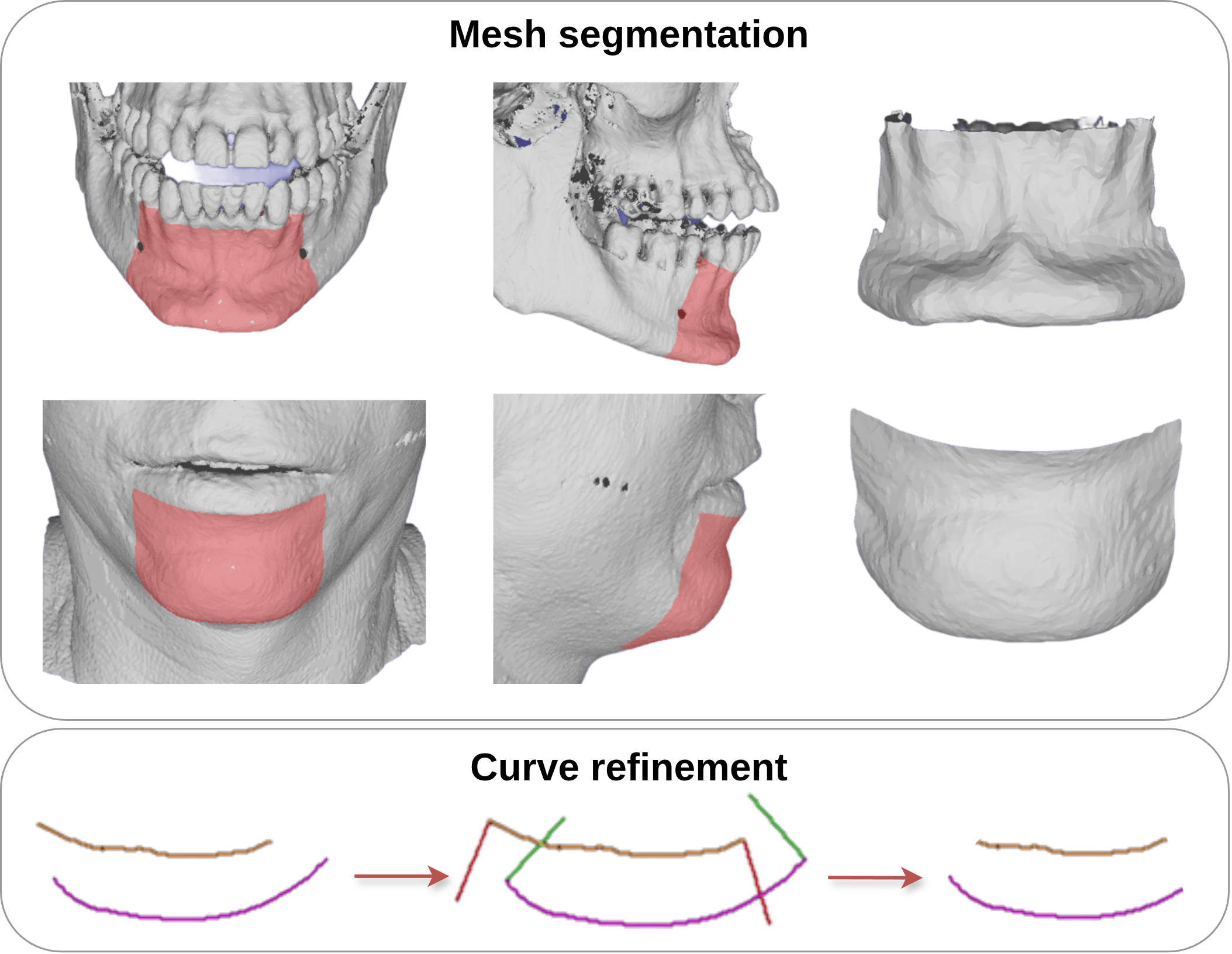}
        \caption{
        Contour extraction for the parallelism term \(\text{P}_{\text{pll}}\). 
        The regions of interest are isolated from the skull and the facial meshes using anatomically defined cutting planes (top).
        The boundary of each segmented region is extracted, cleaned, ordered, and merged into a single continuous curve. The curves are then trimmed using averaged endpoint normals to ensure consistent and comparable extents between skull and face (bottom).}
    \label{fig:curves}
\end{figure}

Finally, the $\text{P}_{\text{pll}}$ term aims to enforce morphological consistency by evaluating the alignment between the projected cranial region and the corresponding facial region in the image. In our proposal, the parallelism metric is applied according to the facial pose in the image: for frontal poses, the chin and jaw regions are analyzed, while for lateral poses the forehead region is examined. For that purpose, we first extract the relevant contour curves via a three‐stage pipeline:

\begin{enumerate}
    \item Mesh Segmentation. The region of interest is isolated. For the forehead region, we rely on two anatomical landmarks: glabella and metopion    \cite{caple2016standardized}. We use metopion as a reference to define and refine the forehead curve due to its standardized anatomical basis, despite some location imprecision. 
    To extract the region, we place two cutting planes (both parallel to the Frankfurt plane \cite{caple2016standardized}) so that one plane passes through glabella and the other through metopion. Everything between these planes is automatically cropped as the forehead mesh. 
    For the chin-jaw region, a similar approach is applied using anatomical references. Segmentation is performed by placing one horizontal cutting plane through the infradentale and two vertical planes through the mental foramina for skull meshes, or through the cheilion landmarks for facial meshes (see the mesh segmentation panel in Figure~\ref{fig:curves}).

   \item Curve Detection. The boundary of the region is extracted from the segmented mesh by projecting and rendering it, and then applying standard image‐processing techniques to isolate the contour. In the case of the forehead, the raw curve is further constrained by discarding all points outside the vertical span defined by the $y$‐coordinates of the glabella and metopion landmarks.

   \item Curve Refinement. The detected contour is then subjected to a cleaning process, whereby isolated pixels are removed, and the remaining ones are sorted.
   Any disconnected fragments are joined by straight segments to yield a single continuous path. Finally, to make skull and face curves directly comparable, each is trimmed at the point where the averaged normal (computed over the terminal 20\% of its length) intersects its counterpart (see the curve refinement panel in Figure~\ref{fig:curves}).
\end{enumerate}

Using the refined curves, the parallelism metric is computed. First, for each pixel on the cranial curve, the closest pixel on the facial curve is found, and vice versa for any unpaired facial pixels. Then, the distances between the paired points are recorded. Let \(\{(s_i,p_i,d_i)\}_{i=1}^N\) be the list of matched skull‐curve points \(s_i\) and face‐curve points \(p_i\), with Euclidean distance \(d_i=\|s_i-p_i\|\). The parallelism metric is thus 
computed as
\(
  \text{P}_{\text{pll}} = \Delta d + \text{P}_{\rm conv} + \text{P}_{\rm int},
\)
\noindent where:
\begin{itemize}
    \item \(\Delta d\) is the base parallelism metric, which represents the spread of the distances:
    \(
        \Delta d = \max d_i-\min d_i
    \).
    \item \(\text{P}_{\rm conv}\) is a convergence penalty applied when the distances between the two curves at their endpoints deviate significantly from the overall mean. 
    Let
    \(
      \bar d = \frac{1}{N}\sum_{i=1}^{N} d_i
    \)
    be the mean of all pairwise distances. Denote by $d_1$ and $d_N$ the distances between the first matched pair $(s_1,p_1)$ and the last matched pair $(s_N,p_N)$, respectively, and define
    \(
      \delta_1 = \bigl|d_1 - \bar d\bigr|\text{ and }
      \delta_N = \bigl|d_N - \bar d\bigr|
    \).
    If either $\delta_e > 0.25\,\bar d$ where \(e\in\{1,N\}\), a penalty is incurred according to
    \[
      \text{P}_{\rm conv} =
      \begin{cases}
        2\,(\delta_1 + \delta_N) & \text{if both ends converge,}\\
        4\,\delta_e              & \text{if only one end converges,}\\
        0                         & \text{otherwise}
      \end{cases}
    \]
    \item \(\text{P}_{\rm int}\) is an intersection penalty that quantifies any contour‐crossing between skull and face. After rasterizing both curves, the skull pixels lying on the \textit{wrong side} of the face boundary are identified. Each violation contributes a fixed penalty of 1,000 to the total score. In frontal facial poses, a violation occurs whenever a paired point satisfies \(s_i.y \ge p_i.y\), meaning the skull jaw contour lies below the facial chin contour. In lateral poses, the condition depends on the viewing direction:
    \[
      \begin{cases}
        s_i.x \le p_i.x & \text{if the subject is looking left,}\\
        s_i.x \ge p_i.x & \text{if the subject is looking right}
      \end{cases}
    \]
    The total directional penalty is proportional to the number of such violations and is expressed as
    \(
    \text{P}_{\mathrm{int}} = 1{,}000 \sum_{i=1}^{N} \text{penalized}(s_i, p_i),\text{ where}
    \)
    \[
    \text{penalized}(s_i, p_i) =
    \begin{cases}
    1 & \text{if } (s_i, p_i) \text{ violates the condition,} \\
    0 & \text{otherwise.}
    \end{cases}
    \]

\end{itemize}

A low \(\text{P}_{\text{pll}}\) thus indicates anatomical alignment between skull and face in the region of interest.

\section{Experimental Study}
\label{sec:exp}


\subsection{Dataset}
\label{sec:data_generation}

\begin{figure*}[h]
    \begin{center}
        \includegraphics[width=\textwidth]{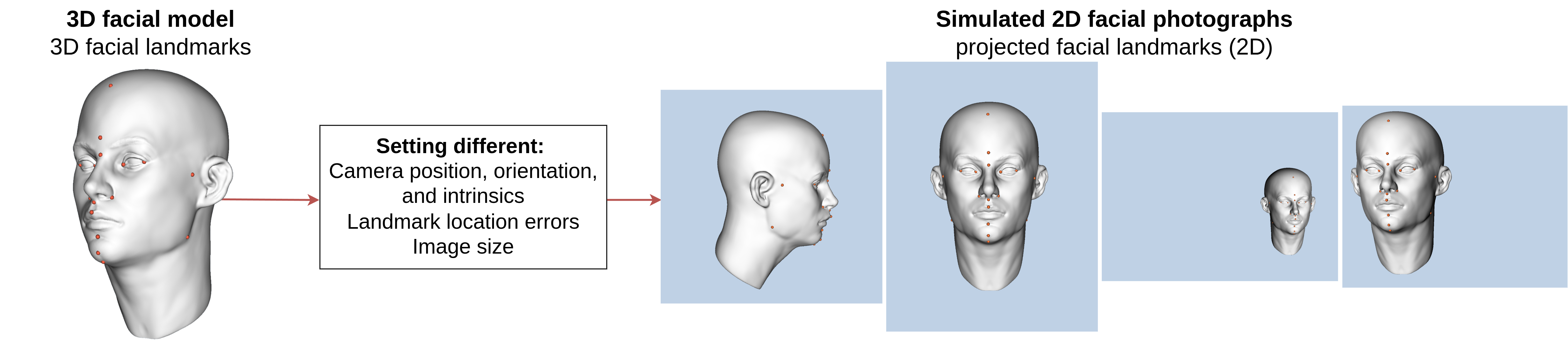}
        \caption{Real facial photographs are replaced by synthetic renderings generated from 3D facial models, where landmarks are precisely defined. These 3D landmarks are projected into 2D using known camera parameters, eliminating annotation errors. This visualization is based on a synthetic face model used solely for illustrative purposes. The underlying 3D model is ``Female head sculpt'' by riceart (\url{https://skfb.ly/I6rZ}), licensed under Creative Commons Attribution 4.0 (\url{https://creativecommons.org/licenses/by/4.0/}).}
	\label{fig:data_generation}
    \end{center}
\end{figure*}

Unlike many established forensic methods, the CFS approach does not rely on ground-truth data, that is, there is no infallible procedure that can yield a perfect, indisputable SFO \cite{DamasHandbook2020}. Moreover, real forensic data are often incomplete and scarce, limiting their availability for research purposes. To overcome these challenges, controlled environments can be utilized to gather data and create synthetic identification scenarios. Building on the framework introduced in \cite{VALSECCHI2018}, we replace real facial photographs with renderings generated from 3D facial models. 
The process is illustrated in~Figure~\ref{fig:data_generation}.

"Female head sculpt." (https://skfb.ly/I6rZ) by riceart is licensed under Creative Commons Attribution (http://creativecommons.org/licenses/by/4.0/).

Our objective is to produce fully specified, error-controlled identification cases in which its corresponding ground-truth solution is known exactly. For each case, we begin by extracting from the CT scan: the 3D skull mesh; the coordinates of cranial landmarks; and the associated soft-tissue vectors, which together define the 3D facial landmark positions (see Section~\ref{sec:cones}). Then, rather than acquiring a conventional photograph and manually annotating 2D facial landmarks (\({f}\)), we render a synthetic image from a 3D facial model. In this model, the 3D facial landmarks are already precisely defined. By applying a known projection \({P}\) (determined by the position, orientation, and intrinsic parameters of the camera) we ensure that
\(
{f_i} = {P}({C_i+\vec{ST_i}}) = {P}({F_i}),
\)
thereby removing all subjectivity and 2D landmark-location error. This rendering approach permits unlimited variation in subject pose and camera parameters, enabling the creation of large, diverse datasets for comprehensive algorithm testing. 

To emulate landmark-location variability, we introduce small, random perturbations \(E_i\) to each ideal 2D landmark:
\(
f_i' = {f}_i + E_i,
\)
which mimics the typical inaccuracies of manual annotation \cite{Campomanes15IJLM}. 
Finally, as part of this data generation process, the facial chin-jaw and forehead regions are segmented and their contour curves are automatically detected and incorporated into the dataset. The methodology is equivalent to that delineated in Section~\ref{sec:chin_forehead}.

Our study involves a total of 17 CT scans (10 male and 7 female subjects) obtained from the New Mexico Decedent Image Database (NMDID) \cite{Edgar2020NMDB}. The ages of the subjects ranged from 20 to 88 years, with an average age of 42 years.
For each subject, we generated 30 frontal and 30 lateral SFO images, resulting in a total of 17,340 superimpositions. This dataset significantly surpasses those used in previous experimental studies on automatic SFO \cite{Ibanez11TFS, Campomanes15IJLM, Ibanez15IJLM, VALSECCHI2018, IBANEZ20121459}. Key parameters for each generated image, including the pose of the camera, focal length, and image size, were randomly selected to reflect a diverse range of realistic scenarios. The choice of landmarks varied across subjects, as some 3D skull models lacked sufficient detail in specific areas. Landmarks that were not visible due to the pose of the subject were excluded from the SFO process. 
The cranial landmarks involved were: left/right (L/R) ectoconchion, L/R gonion, prosthion, glabella, bregma, metopion, L/R dacryon, L/R zygion, L/R alare, L/R frontotemporale, nasion, pogonion, gnathion, and subspinale. The facial ones were: L/R exocanthion, L/R gonion, labiale superius, L/R endocanthion, L/R zygion, L/R alare, L/R frontotemporale, nasion, pogonion, gnathion, subnasale, bregma, glabella, and~metopion.

\subsection{Experimental Settings}
\label{sec:settings}

In order to systematically assess the robustness and effectiveness of our approach, we compared it against the state-of-the-art POSEST-SFO algorithm across three experiments of increasing complexity (A--C). 
The base configuration of our method, Lilium, relies on two fundamental components: the $\text{MSE}_{\text{pix}}$ and the $\text{P}_{\text{cam}}$ terms. From this baseline, we progressively added $\text{P}_{\text{skof}}$ and $\text{P}_{\text{pll}}$ constraints, both separately and in combination, to evaluate the specific contribution of each proposed criterion. 
%
The experiments are defined by the complexity of the scenario:
\begin{itemize}
    \item Exp.~A (Ideal Scenario): No noise was applied to either the 2D facial landmarks or the soft-tissue vectors.
    \item Exp.~B (Landmark Noise): Random noise of up to
    $\pm5$ pixels was added to 2D landmark locations to simulate typical localization errors\cite{Cummaudo13IJLM, Campomanes15IJLM}.
    \item Exp.~C (Realistic Scenario): Soft-tissue vector directions were perturbed by up to $30^\circ$ to simulate estimation uncertainty. This perturbation magnitude represents a deliberately challenging condition, designed to evaluate the model’s robustness under substantial directional noise. In addition, landmark noise was applied concurrently to approximate the most realistic scenario.
\end{itemize}

Since our synthetic data provides known ground-truth values for camera parameters and head pose, we employed them to define acceptance intervals with confidence: $\pm5\%$ for $f_x$, $\pm10\%$ for SCD, and a maximum deviation of $10^\circ$ for head pose (see Section~\ref{sec:camera_params}).

We aim to evaluate identification accuracy using a ranking-based metric \cite{VALSECCHI2018}. For a given facial photograph, a superimposition is performed with every skull in our database. The resulting SFOs are then sorted based on their error metric: the mean pixel-wise projection error for POSEST-SFO and the fitness value for Lilium. The rank of the true matching skull serves as our key performance indicator; a lower rank signifies a higher identification capacity. 
Moreover, we also report the back-projection error (BPE), which is computed as the mean 3D point-to-line distance in mm between each ground-truth 3D facial landmark and the back-projection ray induced by the estimated projection \cite{VALSECCHI2018}. Facial landmarks not visible in the image due to pose are excluded from estimation but included in the BPE computation for evaluation. Lower BPE values indicate a smaller mean 3D displacement of the facial landmarks caused by using the estimated projection instead of the ground-truth projection.

To account for the stochastic nature of DE, all experiments were repeated five times. In Exp.~C, an additional source of variability is introduced through the use of five distinct sets of perturbed soft-tissue vectors, resulting in a total of 25 executions. For every experiment, reported results correspond to the mean ranks and BPE values computed over all associated executions. By contrast, POSEST-SFO is a deterministic method and was therefore executed only once per input instance.

Finally, we assess the quality of the resulting SFOs by means of a metric derived from the \textit{skull-outside-face} criterion described in Section~\ref{sec:skull_face}.
We repurposed $\text{P}_{\text{skof}}$ as an a posteriori evaluation measure to enable a fair comparison between different SFO methods. 
The quality score is defined as the percentage of SFOs in which any of the skull pixels do not overlap onto a pixel of the facial mask. 
Lower percentages indicate higher-quality, anatomically consistent overlays, whereas higher values correspond to poorer alignments. Since several methods are evaluated under multiple runs or across different datasets, we report worst-case performance in terms of the percentage of implausible SFOs to explicitly account for variability and to assess robustness under adverse conditions.


All experiments were conducted on a supercomputer delivering 4.36 PetaFLOPS across 357 nodes, totaling 22,848 Intel Xeon Ice Lake 8352Y cores.

\subsection{Results and Discussion}
\label{sec:results}

Table~\ref{tab:merged_experiment_results_fitness} reports the mean frontal and lateral ranks and BPE values, along with runtime, for all methods across three experimental scenarios. Table~\ref{tab:posest_vs_skullchin} reports the statistical significance of the differences between POSEST-SFO and the Lilium variant with the most competitive mean rank or BPE.

\begin{table*}[h]
    \centering
    \caption{Mean rank, BPE, and computational time results from Exps. A--C. Bold indicates the best value in each column.}
    \label{tab:merged_experiment_results_fitness}
    \footnotesize
    \renewcommand{\arraystretch}{1.1}
    \setlength{\tabcolsep}{3pt}
    \begin{tabular}{l|cc|cc|c|cc|cc|c|cc|cc|c}
        \toprule
        & \multicolumn{5}{c|}{\textbf{Exp A (Ideal)}} & 
          \multicolumn{5}{c|}{\textbf{Exp B (Landm. Noise)}} & 
          \multicolumn{5}{c}{\textbf{Exp C (Realistic)}} \\
        \cmidrule(lr){2-6} \cmidrule(lr){7-11} \cmidrule(lr){12-16}
        & \multicolumn{2}{c|}{\textbf{Rank}} & \multicolumn{2}{c|}{\textbf{BPE}} & \textbf{Time} & 
          \multicolumn{2}{c|}{\textbf{Rank}} & \multicolumn{2}{c|}{\textbf{BPE}} & \textbf{Time} & 
          \multicolumn{2}{c|}{\textbf{Rank}} & \multicolumn{2}{c|}{\textbf{BPE}} & \textbf{Time} \\
        \cmidrule(lr){2-3} \cmidrule(lr){4-5} \cmidrule(lr){6-6} 
        \cmidrule(lr){7-8} \cmidrule(lr){9-10} \cmidrule(lr){11-11} 
        \cmidrule(lr){12-13} \cmidrule(lr){14-15} \cmidrule(lr){16-16}
        \textbf{Method} & Frontal & Lateral & Frontal & Lateral & (s) & Frontal & Lateral & Frontal & Lateral & (s) & Frontal & Lateral & Frontal & Lateral & (s) \\
        \midrule
        POSEST-SFO & \textbf{1.18} & 1.135 & \textbf{1.078} & 2.158 & \textbf{0.014} & \textbf{1.502} & 1.424 & 1.861 & 2.998 & \textbf{0.018} & 2.147 & 2.094 & 2.328 & 4.103 & \textbf{0.017} \\
        Lilium   & 1.552 & 1.883 & 1.15 & 1.998 & 220.286 & 1.672 & 1.626 & 2.296 & 2.769 & 222.439 & 1.909 & 1.781 & 2.608 & 3.311 & 223.757 \\
        Lilium ($\text{P}_{\text{pll}}$) & 1.632 & 1.259 & 1.229 & 2.223 & 515.921 & 1.693 & 1.257 & 1.811 & 2.599 & 515.432 & 1.95 & 1.374 & 2.174 & 2.92 & 512.609 \\
        Lilium ($\text{P}_{\text{skof}}$)  & 1.412 & 1.34 & 1.084 & \textbf{1.894} & 386.235 & 1.625 & 1.271 & 2.07 & 2.539 & 387.511 & 1.899 & 1.341 & 2.299 & 2.87 & 385.442 \\
        Lilium ($\text{P}_{\text{skof}}$+$\text{P}_{\text{pll}}$) & 1.451 & \textbf{1.1} & 1.132 & 2.099 & 537.961 & 1.576 & \textbf{1.111} & \textbf{1.679} & \textbf{2.499} & 536.965 & \textbf{1.897} & \textbf{1.216} & \textbf{1.967} & \textbf{2.748} & 535.077 \\
        \bottomrule
    \end{tabular}
\end{table*}

In ideal conditions (Exp.~A), POSEST-SFO obtains the lowest mean frontal rank (1.18) and the lowest frontal error (1.078\,mm). 
Among the remaining methods, Lilium $\text{P}_{\text{skof}}$ follows, with a mean rank of 1.412 and a comparable frontal BPE (1.084\,mm). 
The Wilcoxon signed-rank test failed to detect a difference in frontal BPE between Lilium $\text{P}_{\text{skof}}$ and POSEST-SFO ($p > 0.999$), suggesting that enforcing skull containment alone is sufficient to approach the geometric accuracy of the state-of-the-art method under ideal conditions.
In lateral alignment, Lilium $\text{P}_{\text{skof}}$+$\text{P}_{\text{pll}}$ outperforms POSEST-SFO in terms of rank (1.1 vs 1.135), although the Wilcoxon signed-rank test failed to detect a difference ($p = 0.201$), while exhibiting a marginally lower BPE (2.099\,mm vs 2.158\,mm), indicating comparable 3D alignment accuracy.
Notably, Lilium $\text{P}_{\text{skof}}$ achieves the lowest lateral BPE overall (1.894\,mm). 
%

Under noisy or perturbed conditions (Exps.~B and C), the advantage of combining $\text{P}_{\text{skof}}$ and $\text{P}_{\text{pll}}$ in Lilium becomes evident. In Exp.~B, for frontal alignments, Lilium $\text{P}_{\text{skof}}+\text{P}_{\text{pll}}$ achieves a mean rank very close to POSEST-SFO (1.576 vs.\ 1.502), with no statistically significant differences detected ($p > 0.999$), while exhibiting a lower frontal BPE (1.679\,mm vs 1.861\,mm, $p < 0.0001$), indicating that anatomical constraints improve geometric accuracy even under landmark noise. In lateral poses, the fully constrained Lilium achieves a significantly lower rank (1.111 vs 1.424, $p < 0.0001$) and BPE (2.499\,mm vs 2.998\,mm, $p = 0.0001$) than POSEST-SFO, demonstrating that the combination of 
priors stabilizes optimization under noisy conditions. In Exp.~C (realistic scenario), Lilium $\text{P}_{\text{skof}}+\text{P}_{\text{pll}}$ significantly outperforms POSEST-SFO in both frontal (1.897 vs 2.147, $p < 0.0001$) and lateral (1.216 vs 2.094, $p < 0.0001$) poses, with lower BPEs (1.967 vs 2.328\,mm frontal, 2.748 vs 4.103\,mm lateral, both $p < 0.0001$). 

A consistent trend observed across all experiments was the superior identification performance of lateral poses compared to frontal ones. Lateral views offer a more constrained and discernible profile of features like the jawline, forehead, and nose, providing less ambiguous cues for alignment. Frontal poses, conversely, are more challenging due to the potential for bilateral landmarks to compensate for each other and the risk of landmark coplanarity, which can lead to multiple 3D poses producing the same 2D projection.
Note that the BPE values show the opposite tendency, being consistently smaller in frontal views. This is because in frontal views, most facial landmarks are fully visible, facing the camera directly. In lateral poses, by contrast, some landmarks are occluded, reducing the effective information available for accurately recovering the 3D configuration.

The superior performance of POSEST-SFO in frontal poses, particularly in Exp.~A and to some extent in Exp.~B, can be attributed to the fact that it relies on ground-truth soft-tissue directional information. 
Under noise-free or low-noise conditions, this access to precise orientation data effectively acts as an oracle, giving POSEST-SFO an advantage that is not available in realistic forensic casework.
In contrast, Lilium leverages anatomical priors to provide robust guidance, allowing it to maintain strong identification performance even in challenging scenarios.


\begin{table}[h]
\centering
\caption{Statistical comparison across frontal and lateral poses (Exps.~A--C).
Significance assessed via Wilcoxon signed-rank test.
Bold indicates $p<0.05$ and better (lower) values.}
\label{tab:posest_vs_skullchin}
\footnotesize
\renewcommand{\arraystretch}{1.05}
\setlength{\tabcolsep}{2.5pt}
\begin{tabular}{llccc|ccc}
\toprule
& & \multicolumn{3}{c|}{\textbf{Rank}} 
& \multicolumn{3}{c}{\textbf{BPE (mm)}} \\
\cmidrule(lr){3-5} \cmidrule(lr){6-8}
\textbf{Exp.} &
\textbf{Pose} &
\textbf{Lilium} &
\textbf{POSEST} &
\textbf{p} &
\textbf{Lilium} &
\textbf{POSEST} &
\textbf{p} \\
\midrule
\multirow{2}{*}{A} & Frontal  & 1.412 & \textbf{1.18} & \textbf{$<\text{0.0001}$}
             & 1.084 & \textbf{1.078} & $>0.999$ \\

 & Lateral  & \textbf{1.1} & 1.135 & 0.201
             & \textbf{1.894} & 2.158 & \textbf{0.001} \\
\midrule
\multirow{2}{*}{B} & Frontal  & 1.576 & \textbf{1.502} & $>0.999$
             & \textbf{1.679} & 1.861 & \textbf{$<\text{0.0001}$} \\

 & Lateral  & \textbf{1.111} & 1.424 & \textbf{$<\text{0.0001}$}
             & \textbf{2.499} & 2.998 & \textbf{0.0001} \\
\midrule

\multirow{2}{*}{C} & Frontal  & \textbf{1.897} & 2.147 & \textbf{$<\text{0.0001}$}
             & \textbf{1.967} & 2.328 & \textbf{$<\text{0.0001}$} \\

 & Lateral  & \textbf{1.216} & 2.094 & \textbf{$<\text{0.0001}$}
             & \textbf{2.748} & 4.103 & \textbf{$<\text{0.0001}$} \\
\bottomrule
\end{tabular}
\end{table}

Regarding the worst-case SFO quality, Table~\ref{tab:skull_outside_worst} reports the results across experiments. In Exp.~A, Lilium $\text{P}_{\text{skof}}$ produces anatomically implausible SFOs in 19.47\% of cases, with only 3.03\% of skull pixels lying outside the facial region. Under noisier and more realistic conditions (Exps.~B and C), Lilium $\text{P}_{\text{skof}}+\text{P}_{\text{pll}}$ further improves robustness, limiting implausible overlays to 17.26–19.83\% of cases with similarly low outside-pixel ratios (2.84–3.18\%). 
Note that 
the number of positive SFOs (the skull and face originate from the same subject) classified as implausible remains minimal across all experiments. This indicates that positive pairs are rarely rejected due to anatomical violations. Furthermore, manual inspection reveals that many of the few Lilium implausible cases are
typically due to open-mouth expressions or minor mesh cropping below the chin, rather than genuine anatomical inconsistencies.

\begin{figure*}
    \centering
    \includegraphics[width=\textwidth]{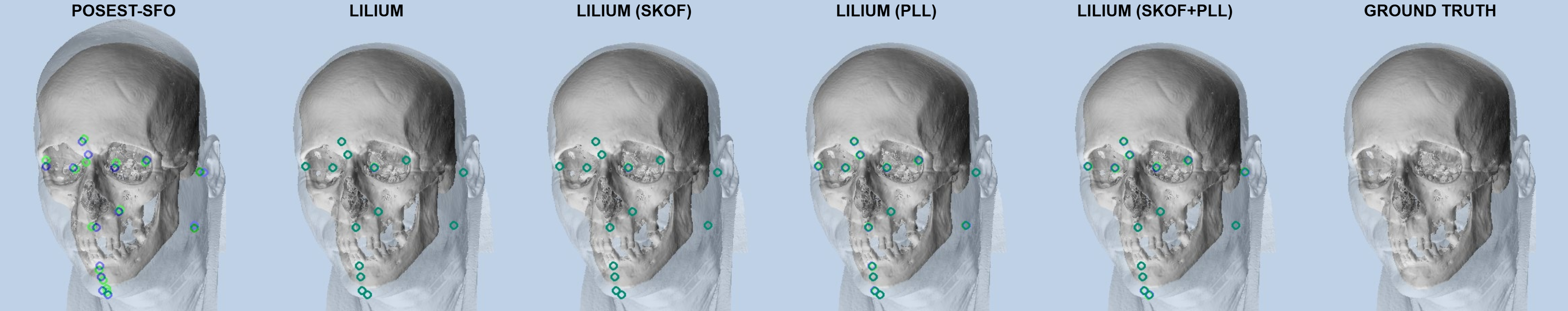}
    \includegraphics[width=\textwidth]{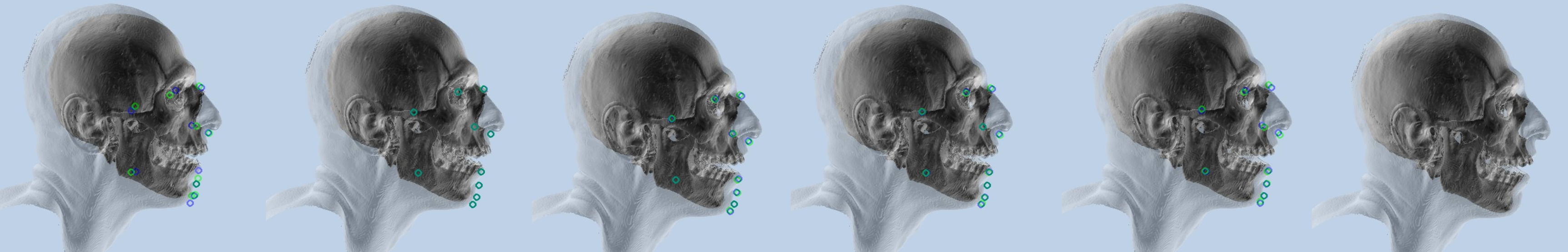}
    \caption{Representative SFO results from Exp.~C across models for multiple subjects. Each column shows a different model; the last is ground truth. Blue circles mark facial landmarks in the image; green dots are projected 3D estimates.}
    \label{fig:qualitative_img}
\end{figure*}

In contrast, POSEST-SFO exhibits a substantially higher incidence of anatomically implausible overlays across all experiments, affecting 70.51–78.29\% of cases and showing larger proportions of skull pixels outside the facial region (3.98–5.08\%). These results demonstrate that low mean-rank or BPE performance alone does not guarantee anatomically realistic SFOs and highlight the importance of explicitly incorporating anatomical plausibility criteria.
Qualitative evaluation of SFOs from the most challenging scenario (Exp.~C, Figure~\ref{fig:qualitative_img}) confirms these trends. POSEST-SFO results show visible misalignment and anatomical inconsistencies, whereas Lilium $\text{P}_{\text{skof}}+\text{P}_{\text{pll}}$ generates overlays that maintain correct landmark positions while respecting craniofacial morphology.

\begin{table}[ht]
\caption{Worst-case implausible SFO metrics across experiments. 
Reported counts correspond to the number of implausible SFOs out of 17{,}340 cases. Bold indicates better (lower) values.}
\label{tab:skull_outside_worst}
\centering
\footnotesize
\setlength{\tabcolsep}{3pt}
\renewcommand{\arraystretch}{1.1}
\begin{tabular}{l l c c c c}
\toprule
\textbf{Exp.} &
\textbf{Method} &
\makecell{\textbf{Implausible} \\ \textbf{SFOs} (\%)} &
\makecell{\textbf{Avg pixels} \\ \textbf{outside} (\%)} &
\makecell{\textbf{Positive} \\ (n)} &
\makecell{\textbf{Negative} \\ (n)} \\
\midrule
\multirow{2}{*}{A}
 & POSEST-SFO & 70.51 & 3.98 & 320 & 11,902 \\
 & Lilium $\text{P}_{\text{skof}}$ & \textbf{19.47} & \textbf{3.03} & \textbf{18}  & \textbf{3,354} \\
\midrule
\multirow{2}{*}{B}
 & POSEST-SFO & 72.6 & 4.51 & 460 & 12,122 \\
 & Lilium $\text{P}_{\text{skof}}+\text{P}_{\text{pll}}$ & \textbf{17.26} & \textbf{3.18} & \textbf{37}  & \textbf{2,952} \\
\midrule
\multirow{2}{*}{C}
 & POSEST-SFO & 78.29 & 5.08 & 634 & 12,937 \\
 & Lilium $\text{P}_{\text{skof}}+\text{P}_{\text{pll}}$ & \textbf{19.83} & \textbf{2.84} & \textbf{48}  & \textbf{3,387} \\
\bottomrule
\end{tabular}
\end{table}


Regarding computational efficiency, POSEST-SFO is extremely fast (under 0.018~s per SFO). Lilium’s runtime scales with optimization complexity, reaching several minutes per SFO for the fully constrained configuration. For forensic applications, this additional computation is justified, as manual superimposition can take hours, while Lilium provides both automated support for identification and anatomically plausible results.


\section{Conclusion}
\label{sec:conclusion}

The SFO stage within CFS presents significant challenges, primarily due to the complexities of aligning a 3D skull with a 2D facial photograph while accounting for soft-tissue uncertainty. This work has introduced Lilium, an innovative automated evolutionary method designed to address this challenge systematically. Lilium explicitly models soft-tissue uncertainty through a novel 3D cone-based representation, with its parameters optimized via a DE algorithm. 

A key strength of our method is its comprehensive fitness function, which guides the optimization process by minimizing landmark projection error while simultaneously penalizing solutions that violate crucial plausibility constraints: (1) unrealistic camera settings ($f_x$ and SCD); (2) inconsistencies between the estimated and image-inferred head poses; (3) SFOs where the projected skull extends beyond the facial boundaries ($\text{P}_{\text{skof}}$ term); and (4) lack of parallelism between skull and facial regions ($\text{P}_{\text{pll}}$ term). Furthermore, Lilium promotes anatomical accuracy by employing a hybrid strategy for the joint optimization of bilateral landmarks, encouraging symmetry while preserving realistic anatomical variation. These innovations enable the generation of robust and accurate overlays making the automated SFO process more analogous to the work conducted by forensic anthropologists.

To validate the proposed methodology, we compared it against the state-of-the-art POSEST-SFO algorithm through three experiments. Each experiment comprised 17,340 superimpositions per run and introduced a distinct noise profile to simulate varying levels of forensic complexity. 
Across the three scenarios, Lilium with combined constraints consistently achieved robust performance. In the ideal, noise-free frontal setting (Exp.~A), POSEST-SFO attained lower ranks. 
Nonetheless, the Wilcoxon signed-rank test failed to detect a difference in BPE in frontal scenarios between POSEST-SFO and Lilium $\text{P}_{\text{skof}}$, indicating comparable geometric accuracy.
Under noisy or perturbed conditions (Exps.~B and C), Lilium $\text{P}_{\text{skof}}+\text{P}_{\text{pll}}$ demonstrated superior robustness. In Exp.~B, although POSEST-SFO obtained a slightly lower mean rank in frontal poses (1.502 vs 1.576), the Wilcoxon signed-rank test failed to detect a difference, whereas Lilium achieved significantly lower BPE values. Regarding lateral poses, Lilium significantly outperformed POSEST-SFO in both rank (1.111 vs 1.424) and BPE (2.499\,mm vs 2.998\,mm). In the most realistic scenario (Exp.~C), Lilium clearly outperformed POSEST-SFO in both frontal (1.897 vs 2.147) and lateral (1.216 vs 2.094) ranks, accompanied by significantly lower BPEs in both poses (frontal: 1.967\,mm vs 2.328\,mm; lateral: 2.748\,mm vs 4.103\,mm).

Regarding the SFO quality evaluation, Lilium $\text{P}_{\text{skof}}$+$\text{P}_{\text{pll}}$ and Lilium $\text{P}_{\text{skof}}$ ensured anatomical plausibility, reducing worst-case implausible overlays from 70.51–78.29\% with POSEST-SFO to only 17.26–19.83\% of cases, while also limiting skull pixels outside the face to 2.84–3.18\%. Qualitative results from Exp.~C confirmed that fully constrained Lilium generates visually coherent and anatomically consistent overlays, balancing identification accuracy with morphological fidelity. Although POSEST-SFO offers faster execution times, this comes at the expense of anatomical reliability. Lilium requires higher computational costs, but remains acceptable in forensic practice where the manual superimposition process typically takes several hours.

In summary, this study demonstrates that leveraging anatomical and morphological criteria substantially enhances SFO performance, offering crucial robustness to noise. 
Future work will focus on several complementary directions. First, we plan to refine the modeling of soft-tissue uncertainty by exploring irregular 3D regions dynamically adapted to direction-specific variability. Second, an important line of future investigation will involve extending validation to larger and more diverse datasets, including real forensic imagery, in order to better characterize these sources of uncertainty. Third, from a computational perspective, future developments will investigate performance optimization strategies, including GPU-accelerated rendering, as well as parallel implementations of the optimization pipeline.
Finally, we intend to integrate additional anatomical constraints to further improve robustness and to support broader acceptance of the method within the forensic science community.

\section*{Acknowledgments}

This publication is part of the R\&D\&I project PID2024-156434NB-I00 (CONFIA2), funded by MICIU/AEI/10.13039/501100011033 and ERDF, EU.
Miss Martínez-Moreno is supported by grant PRE2022-102029 funded by MICIU/AEI/10.13039/501100011033 and by FSE+.
Dr. Valsecchi's work is supported by Red.es under grant Skeleton-ID2.0 (2021/C005/00141299).



This contribution made use of the Free Access Decedent Database, funded by the National Institute of Justice under grant number 2016-DN-BX-0144.
Moreover, the authors would like to thank R. Guerra, M.A. Guativonza, F. Navarro for their valuable contributions and support throughout the development of this work.

Dr. Valsecchi, Dr. Mesejo, Miss Martínez-Moreno, and Prof. Damas among other researchers who did not collaborate in the current study are coinventors with different percentages of ownership of an invention titled ``Method for forensic identification by automatically comparing a 3D model of a skull with one or more photos of a face'', for which a patent application has been filed in the US (No. US 2024/0320911 AI) among other countries.

\bibliography{bibliography.bib}
\bibliographystyle{IEEEtran}


 




\vfill

\end{document}